\documentclass[conference]{IEEEtran}

\usepackage{cite}
\usepackage{amsmath,amssymb,amsfonts}
\usepackage{algorithmic}
\usepackage{graphicx}
\usepackage{textcomp}
\usepackage{xcolor}
\usepackage{multirow}
\usepackage{tabularx}
\usepackage{float}
\usepackage{caption}
\usepackage{stfloats}
\usepackage{amsmath} 
\usepackage{amssymb} 
\usepackage{svg}
\usepackage{comment}

\def\BibTeX{{\rm B\kern-.05em{\sc i\kern-.025em b}\kern-.08em
    T\kern-.1667em\lower.7ex\hbox{E}\kern-.125emX}}
\begin{document}

\title{LogLLaMA: Transformer-based log anomaly detection with LLaMA\\

}

\author{\IEEEauthorblockN{Zhuoyi Yang}
\IEEEauthorblockA{
\textit{University of California, Irvine}\\
Irvine, CA, US \\
zhuoyy1@uci.edu}
\and
\IEEEauthorblockN{Ian G. Harris}
\IEEEauthorblockA{
\textit{University of California, Irvine}\\
Irvine, CA, US \\
harris@ics.uci.edu}
}

\maketitle

\begin{abstract}
Log anomaly detection refers to the task that distinguishes the anomalous log messages from normal log messages. Transformer-based large language models (LLMs) are becoming popular for log anomaly detection because of their superb ability to understand complex and long language patterns. In this paper, we propose LogLLaMA, a novel framework that leverages LLaMA2. LogLLaMA is first finetuned on normal log messages from three large-scale datasets to learn their patterns. After finetuning, the model is capable of generating successive log messages given previous log messages. Our generative model is further trained to identify anomalous log messages using reinforcement learning (RL). The experimental results show that LogLLaMA outperforms the state-of-the-art approaches for anomaly detection on BGL, Thunderbird, and HDFS datasets.
\end{abstract}

\begin{IEEEkeywords}
Anomaly detection, LLMs, RL, Natural Language Processing (NLP) 
\end{IEEEkeywords}

\section{introduction}
System logs, which record detailed system events, serve as a vital source of information for system monitoring, debugging, and security auditing. They offer insights into system performance and potential issues, with anomalies often indicating system faults, security breaches, or operational failures. As computer systems become increasingly vulnerable to cyberattacks, accurate detection of log anomalies is essential to preventing attacks and malfunctions \cite{LogOverview}.

Log anomaly detection frameworks take structured log messages as input. The goal is to classify log messages into normal or abnormal categories with precision. The fundamental approach involves identifying distinct patterns in the logs and differentiating typical behaviors from anomalies.

Because system log files are complex, large, and require automation in identifying patterns and making predictions, many machine learning methods are proposed on anomalous log message classifications. Most of such methods are either unsupervised such as Principal Component Analysis (PCA)\cite{HDFS_dataset&PCA}, iForest \cite{iForest}, and LogCluster \cite{LogCluster}, or semi-supervised such as One-Class Support Vector Machine \cite{OCSVM}. However, traditional machine learning models, which are based on hand-crafted features, are infeasible to capture the temporal information of discrete log messages \cite{LogBert}.

More recently, many deep-learning-based approaches have been proposed to detect anomalous log messages. Most works in this area treat log messages as if they are natural language. For example, DeepLog \cite{DeepLog} uses LSTM models to learn the contextual features in normal log sequences and detect abnormal patterns based on predictions. Experimental results show that these approaches achieve higher accuracy than traditional machine learning approaches at log anomaly detection. However, these models exhibit limitations in effectively capturing long sequential log messages. Identifying an anomalous log message often requires considering a substantial number of preceding messages but LSTM-based models often are constrained in their ability to process sequences of the necessary length.

With the development of deep learning, transformers \cite{attention_is_all_you_need} have been widely adopted, achieving remarkable success in natural language processing (NLP) tasks through large language models (LLMs) \cite{transformers_in_nlp}. This success has inspired their applications in log anomaly detection, as log messages are often treated as natural language. Among all the LLMs, BERT \cite{BERT} is the most utilized one for log anomaly detection so far. For instance, LogBERT \cite{LogBert} utilizes a BERT-based model to identify anomalies through masked token prediction. While such models inherit the benefits of BERT, they are also affected by the disadvantages of BERT such as its sensitivity to the number of masked tokens and its maximum input length. Therefore, BERT-based models may not fully capture the sequential and temporal relationships between log messages.

We introduce a novel framework LogLLaMA, which is built upon LLaMA2 \cite{llama2openfoundation}. LogLLaMA harnesses the power of the transformer architecture to capture the complex pattern of log messages. Our framework consists of three modules. First, it preprocess the raw log messages by parsing them. Then, the LLaMA2-based foundation model is finetuned with the preprocessed log messages, after which the model is capable of predicting the next log sequences given previous log messages. Finally, we employ reinforcement learning, incorporating the REINFORCE algorithm \cite{REINFORCE} as the update policy. The algorithm is augmented with an entropy bonus, reward clipping, and a customized reward system to effectively identify anomalous log messages.

Our contributions are as follows: 1) We propose a new framework LogLLaMA, which is capable of capturing the underlying pattern of normal log messages. After finetuning, our model has the ability to generate new log key sequences given previous log messages. Built on top of LLaMA2, LogLLaMA closes the gap between log messages and natural language. 2) We incorporate the REINFORCE algorithm with entropy bonus and reward clipping as well as a customized reward system for anomaly detection. With Top-K selection, LogLLaMA is able to identify anomalous log messages with high accuracy. 3) By conducting experiments on three publicly available datasets, we compare our framework with multiple baselines. The results demonstrate that LogLLaMA outperforms state-of-the-art methods by a significant margin.
\section{related work}
Log anomaly detection is critical to finding security leaks and maintain system reliability. Various machine learning methods have been proposed: Principal Component Analysis(PCA)-based algorithms \cite{HDFS_dataset&PCA} are used to separate out repeating patterns in feature vectors, thereby making abnormal message patterns easier to detect. Isolation forest (iForest) \cite{iForest} randomly selects a feature and a split value between its maximum and minimum values. This process creates shorter paths in trees for anomalous data points, distinguishing them from the rest of the data. One-class Support Vector Machines (OCSVM) \cite{OCSVM} operates by defining a boundary around the majority class (normal instances) in the feature space. This boundary is constructed to encapsulate the normal data points, effectively creating a region of normalcy. Although these models are capable of identifying outliers in the log data, a severe issue is that they are trying to extract the semantics of log events through feature engineering, which only focuses on the words in the log event, without considering the interaction between words in the log events, or word order \cite{LogST}. However, log messages are often mixed and interdependent, with individual words frequently lacking standalone significance.

Another category of approaches for log anomaly detection is deep-learning methods. DeepLog \cite{DeepLog} is the first to use LSTM models to model the sequence of log events. The LSTM learns to predict the next log event given the previous sequences. Its prediction ability is based on the temporal patterns of normal log sequences. LogAnomaly extends DeepLog and proposes the Template2Vec approach to extract both sequential and quantitative features \cite{Yan2024}. 
%The fundamental logic of these deep-learning methods is to process log messages as if they were natural languages. 
Previous studies have shown that deep-learning methods have much higher accuracy than traditional machine learning methods \cite{BERT,Yan2024}. Our experimental results also show that deep-learning methods have a more consistent performance across different datasets than traditional machine learning methods. 

More recently, LLMs with transformers \cite{attention_is_all_you_need} have demonstrated their power in different NLP tasks \cite{transformers_in_nlp, Power_of_LLMs}. Although LLMs are not specifically trained to understand logs, they can be finetuned on log datasets to learn the patterns and structures of log messages with the right downstream tasks. LogBERT \cite{LogBert} is the first to use Bidirectional Encoder Representations from Transformers (BERT) as a base model to predict the masked tokens of given log events. LogBERT treats log entries as sequences of events, encoding their semantic and contextual information using BERT's transformer-based architecture. LogBERT is finetuned on normal log sequences to learn their patterns and structures. During inference, LogBERT predicts whether a given log event or sequence deviates from the normal patterns it has learned. LogST\cite{LogST} uses novel log semantics representation method to capture the semantic meaning of the log events and then feed the encoded log representation into the BERT base model. PreLog \cite{PreLog} finetunes the model with contrastive learning that unifies different log analytics tasks into a single framework. The model adopts the sequence-to-sequence transformer architecture and is finetuned on a large amount of unlabeled log data. These models mostly leverage BERT\cite{BERT} and therefore they inherit the benefits of BERT, such as its self-attention mechanism which helps it avoid local bias, and its use of bidirectional context allows it to capture complex and ambiguous contextual relationships \cite{LogFit}. However, they also inherit the disadvantages of BERT. BERT-based models are very sensitive to the ratio of the masked tokens. As a result, extensive experiments are needed to find the right ratio. In addition, BERT itself is typically limited to processing sequences of up to 512 tokens. For tasks involving longer log sequences, information may be lost unless the input is truncated or split, which can lead to suboptimal results. Confined by the input length, BERT-based models may not fully capture the sequential and temporal relationships between log messages. In this work, we propose LogLLaMA, which leverages the LLaMA2 model to understand patterns of normal log messages by training it to predict the next log key, and we implement a new RL mechanism to boost the accuracy of log anomaly detection.
\section{LogLLaMA}

\subsection{Overview of our Framework}

\begin{comment}
\begin{figure*}[ht]
    \centering
    \includegraphics[width=\textwidth]{figures/Framework_Overview.pdf} 
    \vspace{-60pt} % Reduce space by 10pt
    \caption{Framework Overview}
    \label{Framework_overview}
\end{figure*}
\end{comment}

\begin{figure}[ht]    
    \includegraphics[width=\textwidth]{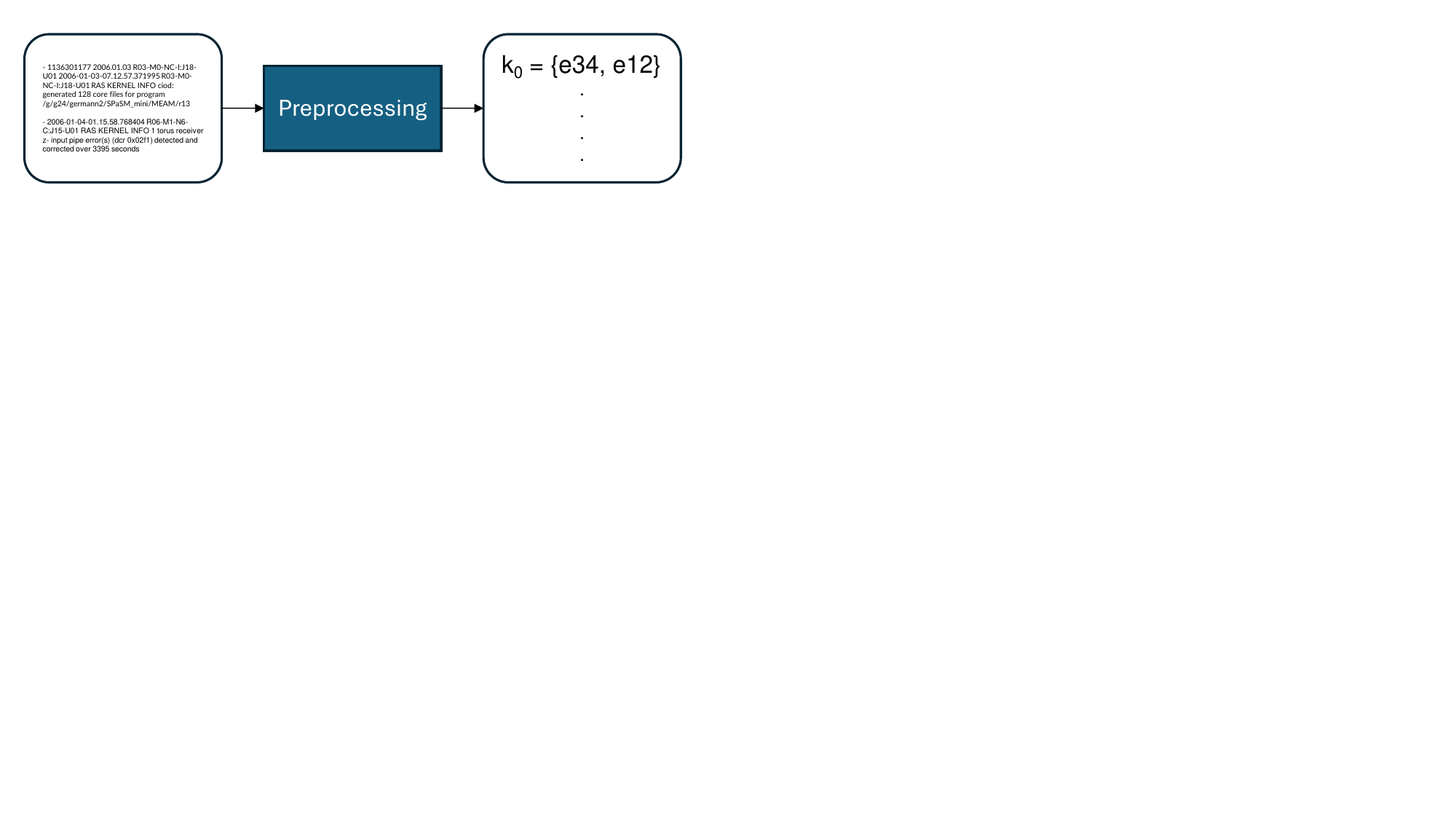} 
    \vspace{-220pt} % Reduce space by 10pt
    \caption{Module 1: Log message preprocessing}
    \label{module1}
\end{figure}

\begin{figure}[ht] 
\centering{
    \includegraphics[width=\columnwidth]{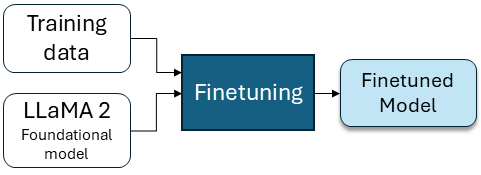} 
%    \vspace{-220pt} % Reduce space by 10pt
    \caption{Module 2: Model finetuning}
    \label{module2}
    }
\end{figure}

\begin{figure}[ht]  
    \centering
    \includegraphics[width=\columnwidth]{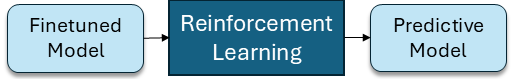} 
%    \vspace{-100pt} % Reduce space by 10pt
    \caption{Module 3: Anomaly detection with RL}
    \label{module3}
\end{figure}

As shown in Figures \ref{module1}, \ref{module2} and \ref{module3}, our framework includes three modules: 1) Log message preprocessing, 2) Model finetuning, and 3) Anomaly detection with RL. In Module 1, log messages are parsed to be log templates and hashed to become compact representations, which we call log keys. Then log keys are grouped to be log key sequences. In Module 2, log key sequences are encoded and used as training data to finetune the model. The model uses the encoding to learn log patterns and gain the ability to generate normal log messages. Finally, in Module 3, the finetuned language model is further trained with RL and becomes a predictive model which can detect anomalous log patterns. 
%Here is a detailed introduction to our framework.

\begin{figure*}[ht]
    \raggedright
    \includegraphics[width=2\columnwidth]{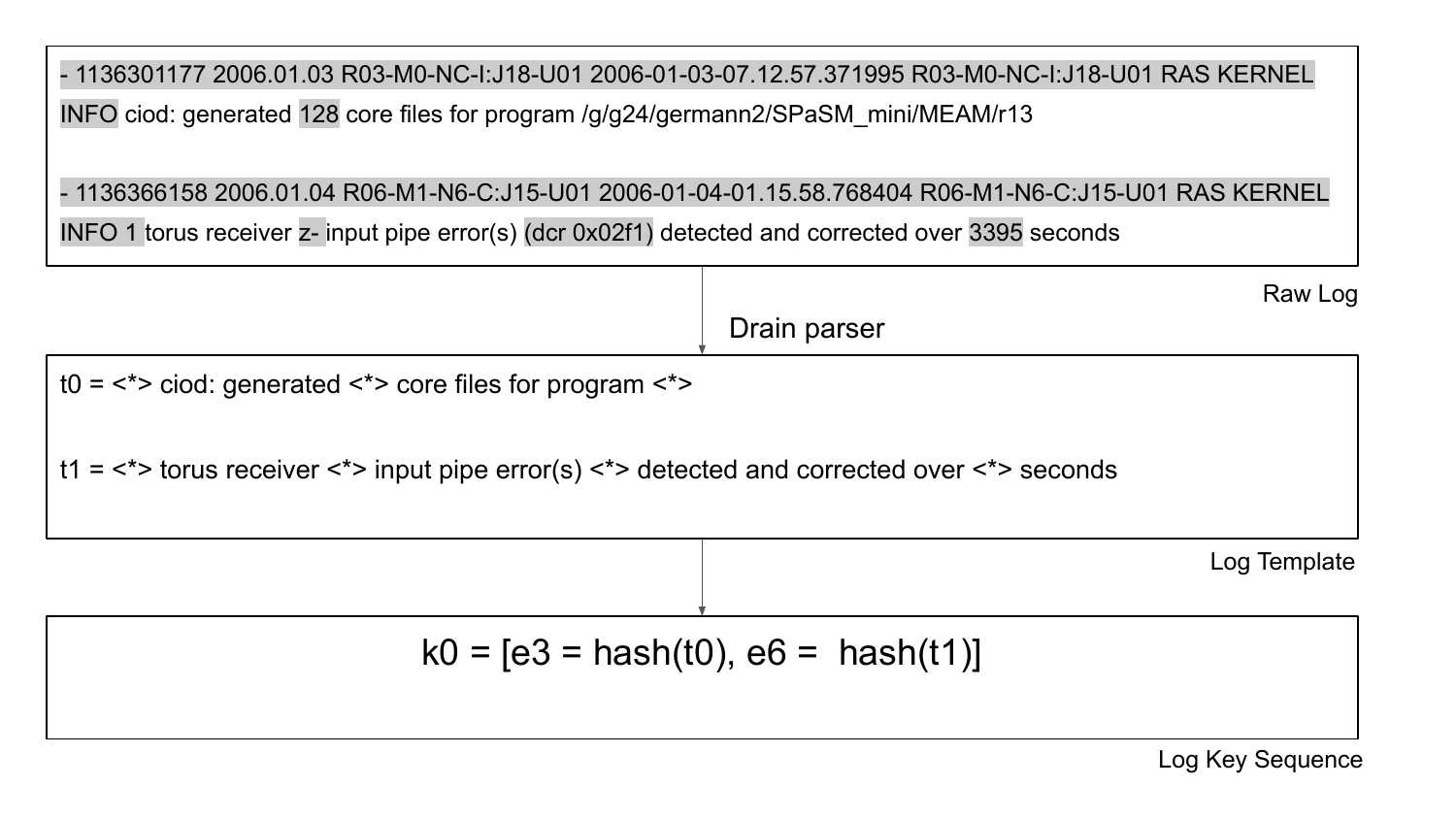} 
    %\vspace{-85pt}
    \captionsetup{justification=raggedright, format=plain}
    \caption{An example of log messages, log templates, and log keys on BGL dataset. The variants are stripped off by the log  parser}  \label{log_preprocessing}
\end{figure*}

\subsection{Module 1: Log Message Preprocessing}
The first step of log anomaly detection is to preprocess the log messages because it is hard to capture the field structure present in most system log messages. Module 1 takes the raw log messages as input and outputs a more structured log representation. Each raw log message consists of two parts: variants and invariants. Variants are the parts that change across different entries including variable values, timestamps, or unique identifiers. The shaded parts in Figure \ref{log_preprocessing} are variants. Invariants refer to the static parts that remain consistent including log level, program path, and message context. Take the following log message from the BGL dataset as an example.
\begin{description}
\item {\bf 1136301177 2006.01.03 R03-M0-NC-I:J18-U01 2006-01-03-07.12.57.371995 R03-M0-NC-I:J18-U01 RAS KERNEL INFO ciod: generated 128 core files for program /g/g24/germann2/SPaSM\_mini/MEAM/r13 }
\end{description}
In this log message, the items "{\bf ciod}", "{\bf generated}", and "core files for program "{\bf /g/g24/germann2/SPaSM\_mini/MEAM/r13}" are considered as invariants, while the remaining items are considered to be variants.

We adopt the Drain algorithm \cite{Drain} as the log parser. The log parser strips off the variants of each raw log message and converts it into a structured log template with the invariants. In the template, the stripped parts are represented by <*>. Therefore, each log template represents a distinct type of raw log message. The parser then hashes the templates into fixed-length compact representations, referred to as log keys. Each log key is denoted as \textit{$e_x$, x = 0, 1, 2 ... n} as shown at the bottom of Figure \ref{log_preprocessing}, with \textit{n} being the max number of unique log keys in the dataset. A series of log keys form a log key sequence. We denote each log key sequence as \textit{{$k_x$, x = 0, 1, 2...n}}, with \textit{n} being the number of total log key sequences in a dataset. The number of log key sequence depends on the slicing of each dataset. For example, messages of the HDFS dataset are grouped by session IDs, then the number of log key sequence is equal to the number of session IDs HDFS dataset has.

\subsection{Module 2: Model Finetuning}
In this module, log key sequences are encoded and fed into the model as input data. We leverage the large language model LLaMA2 and finetune it for log anomaly detection. Because LLaMA2 has already been trained with extensive language corpus, its architecture enables it to capture language patterns effectively, making it useful for generating coherent and contextually relevant text. Built on top of LLaMA2, a generative log language model is finetuned using normal log key sequences of the BGL, Thunderbird, and HDFS datasets. This allows the model to learn the patterns and structures of normal log messages. After training, LogLLaMA is capable of generating normal log key sequences given any input log messages. We denote the loss function used during finetuning of the model as the following:
\begin{equation}
L(\theta) = - \frac{1}{N} \sum_{t=1}^{N} \log p(e_{t+1} \mid k_{0:t}; \theta),
\end{equation}
where $\theta$ is the model parameter, \textit{N} is the number of training examples, $e_{t+1}$ is the next true log key, and $k_{0:t}$ is the previous log sequences. The goal of the finetuning process is to minimize this loss function.

\subsection{Module 3: Anomaly Detection with RL}
\begin{figure}[ht]  
    \centering
    \includegraphics[width=0.5\columnwidth]{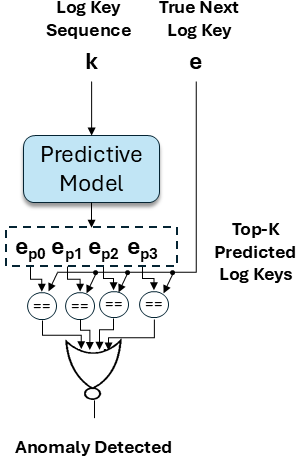} 
    \caption{Top-K Selection}
    \label{fig:mod4}
\end{figure}

In this module, we adopt the REINFORCE algorithm\cite{REINFORCE} with a customized reward system to detect anomalous log messages. As a policy gradient method, REINFORCE is powerful for complex tasks where the optimal actions are not known in advance and must be discovered through experience. It is focused on problems where decisions are made sequentially over time. The agent’s choices influence not only immediate rewards but also future rewards, making it necessary to learn a long-term strategy for maximizing cumulative reward. Instead of comparing a single predicted next log key to the ground truth, we employ a reward mechanism called Top-K selection. As shown in Figure \ref{fig:mod4}, our model outputs a list of K possible sequences as candidates of the next log key. If the true log key is in the list, this log key is normal. Otherwise, if the Top-K prediction list does not include the true log key, this log key is considered anomalous. We define the following elements in our model.

\textbf{Agent.} The agent is our LogLLaMA model, which is generating successive log keys in a sequence based on previous log key sequences. The agent observes the current state $S_t$, samples an action $a_t$ based on policy $\pi_\theta(a_t|S_t)$, receives a reward $r_t$ as feedback from the environment for its action and updates its policy $\pi_\theta$ using the REINFORCE algorithm to maximize the expected cumulative reward.
  
\textbf{State.} The state ${S_{0:t}}$ at each log key position \textit{t} encapsulates all the information the agent uses to make its decision which includes features of the current log message as well as the contextual information from previous log messages. 

\textbf{Action.} The action is to generate the next log keys at each position. The next log key is chosen from the distribution of most possible tokens given by the our model's logits. The model makes this decision based on the previous log keys it has already generated, which is the the previous state up to log key position \textit{t}, $S_{0:t}$. Therefore, the action $a_{t+1}$ at the next position \textit{t} + 1 can be denoted as $a_{t+1} \sim \text{Top-K}(P(e_{t+1}|S_{0:t}))$.

\textbf{Reward.} The reward at position \textit{t} is denoted as $r_t$. It is computed based on the quality of the action. Instead of comparing a single predicted next log key to the ground truth, we employ a reward mechanism called Top-K selection. Our model outputs a list of K possible sequences as candidates of the next log key. If the true log key is in the list, this log key is normal. Otherwise, if the Top-K prediction list does not include the true log key, this log key is considered anomalous. In addition, we also use entropy bonus and reward clipping to increase the stability and flexibility of our model. 

\begin{itemize}
    \item Top-K match: If the true log key is in the Top-K candidate list, the model is considered to have correctly predicted the log key. The agent receives a reward of +1.
    \item Top-K miss: If the true log key is not in the Top-K candidate list, one of the two conditions is true: 1) the true log key is an anomaly so the model can not predict the correct following log keys, or the true log key is normal and model performs poorly. For the first case, the model is doing what it is expected to do, so the agent receives a reward of +1. For the second case, the agent is penalized by a reward of -1. 

    \item Entropy bonus: The entropy bonus adds a small value proportional to the uncertainty of the model's predictions. When the model is confident, its log key prediction distribution will have low entropy. If the entropy is high, it means the model is uncertain and several log keys have similar probabilities. By rewarding higher entropy, the model is encouraged to explore less likely tokens, which leads the model to discovering better solutions during training. The entropy of the token distribution is used to compute this bonus.
    \item Reward clipping: We employ reward clipping to ensure that extremely high or low values of the reward do not cause instability in the training process. If the reward becomes too large or small, it might lead to large gradients that destabilize training. Clipping the rewards ensures that the reward is kept within a safe range, providing more consistent gradients during backpropagation.
\end{itemize}

The value of \textit{K} is a hyper-parameter that controls the strictness of the model. Using Top-K mechanism helps the model avoid being overly deterministic. Sampling only the most probable log key can result in repetitive or overly predictable outputs, while considering the top K introduces more variety. A single most probable value may not always be correct, especially in cases where the probabilities are close or the model is uncertain. By focusing on the top K values, the model incorporates a measure of uncertainty, acknowledging that multiple plausible choices exist. Top-K selection increases robustness by including slightly less probable values, which might still be correct in noisy or ambiguous scenarios.

\textbf{Policy.} The policy is the strategy that the agent uses to decide which actions to take. The policy is implicitly encoded in our model. Our model generates logits which are used by the model to select the next log key through sampling. The agent operates according to a policy, denoted as $\pi_\theta(a_t|S_t)$, which is a probability distribution over possible actions $a_t$, given the current state $S_t$, The policy is parameterized by $\theta$. The agent selects actions probabilistically based on this policy.

\textbf{Policy update.} We use REINFORCE algorithm \cite{REINFORCE} to update the policy after each step. The objective is to maximize the expected cumulative reward $J(\theta)$, defined as: 
\begin{equation}
J(\theta) = \mathbb{E}_{\pi_\theta} \left[ \sum_{t=0}^T r_t \right]. 
\end{equation}
To optimize $J(\theta)$, the policy computes the gradient of the objective function. Using the log-derivative trick, the gradient can be expressed as:
\begin{equation}
\nabla_\theta J(\theta) = \mathbb{E}_{\pi_\theta} \left[ \sum_{t=0}^{T} \nabla_\theta \log \pi_\theta(a_t | s_t) R_t \right],
\end{equation}
where $R_t$ is the cumulative reward starting from position \textit{t}, and \( \nabla_\theta \log \pi_\theta(a_t | s_t) R_t \) is the gradient of the log-probability of the action $a_t$.

The policy is updated on $\theta$ using the following rule:
\begin{equation}
\theta \gets \theta + \alpha \nabla_\theta J(\theta),
\label{eq:update_rule}
\end{equation}
where $\alpha$ is the learning rate.

\begin{table*}[h!]
\centering  
    \caption{Experimental Results of Datasets BGL, HDFS, and Thunderbird Datasets.}
    \label{experimental results for all models}
    \resizebox{\textwidth}{!}{%
    \begin{tabular}{|c|ccc|ccc|ccc|}
    \hline
    \multirow{2}{*}{} &
      \multicolumn{3}{c|}{BGL} &
      \multicolumn{3}{c|}{HDFS} &
      \multicolumn{3}{c|}{Thunderbird} \\ \cline{2-10} 
     &
      \multicolumn{1}{l|}{F1 score} &
      \multicolumn{1}{l|}{Precision} &
      \multicolumn{1}{l|}{Recall} &
      \multicolumn{1}{l|}{F1 score} &
      \multicolumn{1}{l|}{Precision} &
      \multicolumn{1}{l|}{Recall} &
      \multicolumn{1}{l|}{F1 score} &
      \multicolumn{1}{l|}{Precision} &
      \multicolumn{1}{l|}{Recall} \\ \hline
    PCA &
      \multicolumn{1}{c|}{0.051} &
      \multicolumn{1}{c|}{0.211} &
      0.029 &
      \multicolumn{1}{c|}{0.039} &
      \multicolumn{1}{c|}{0.045} &
      0.034 &
      \multicolumn{1}{c|}{0.592} &
      \multicolumn{1}{c|}{0.423} &
       \textbf{0.997}\\ \hline
    iForest &
      \multicolumn{1}{c|}{0.077} &
      \multicolumn{1}{c|}{0.487} &
      0.042 &
      \multicolumn{1}{c|}{0.078} &
      \multicolumn{1}{c|}{0.043} &
      0.422 &
      \multicolumn{1}{c|}{0.043} &
      \multicolumn{1}{c|}{0.289} &
      0.023 \\ \hline
    OCSVM &
      \multicolumn{1}{c|}{0.137} &
      \multicolumn{1}{c|}{0.082} &
      0.421 &
      \multicolumn{1}{c|}{0.062} &
      \multicolumn{1}{c|}{0.032} &
      \textbf{0.904} &
      \multicolumn{1}{c|}{0.659} &
      \multicolumn{1}{c|}{0.499} &
      0.968 \\ \hline
    LogCluster &
      \multicolumn{1}{c|}{0.746} &
      \multicolumn{1}{c|}{\textbf{0.952}} &
      0.613 &
      \multicolumn{1}{c|}{0.458} &
      \multicolumn{1}{c|}{\textbf{0.987}} &
      0.298 &
      \multicolumn{1}{c|}{0.443} &
      \multicolumn{1}{c|}{\textbf{0.967}} &
      0.287 \\ \hline
    DeepLog &
      \multicolumn{1}{c|}{0.870} &
      \multicolumn{1}{c|}{0.801} &
      0.952 &
      \multicolumn{1}{c|}{0.828} &
      \multicolumn{1}{c|}{0.768} &
      0.897 &
      \multicolumn{1}{c|}{0.898} &
      \multicolumn{1}{c|}{0.823} &
      0.987 \\ \hline
    LogAnomaly &
      \multicolumn{1}{c|}{0.867} &
      \multicolumn{1}{c|}{0.884} &
      0.850 &
      \multicolumn{1}{c|}{0.510} &
      \multicolumn{1}{c|}{0.899} &
      0.356 &
      \multicolumn{1}{c|}{0.930} &
      \multicolumn{1}{c|}{0.873} &
      0.996 \\ \hline
    LogBert &
      \multicolumn{1}{c|}{0.878} &
      \multicolumn{1}{c|}{0.874} &
      0.882 &
      \multicolumn{1}{c|}{0.718} &
      \multicolumn{1}{c|}{0.698} &
      0.739 &
      \multicolumn{1}{c|}{0.948} &
      \multicolumn{1}{c|}{0.943} &
      0.954 \\ \hline
    \textbf{LogLLaMA} &
      \multicolumn{1}{c|}{\textbf{0.95927}} &
      \multicolumn{1}{c|}{0.92748} &
      \textbf{0.99333} &
      \multicolumn{1}{c|}{\textbf{0.89363}} &
      \multicolumn{1}{c|}{0.93897} &
      0.85248 &
      \multicolumn{1}{c|}{\textbf{0.97407}} &
      \multicolumn{1}{c|}{0.95679} &
      0.99198 \\ \hline
    \end{tabular}%
    }

\vspace{1cm}
    \centering
    \caption{Performance of LogLLaMA with or without reinforcement
learning.}
    \label{ablation study table}
    \resizebox{\textwidth}{!}{%
   
    \begin{tabular}{|c|ccc|ccc|ccc|}
\hline
\multirow{2}{*}{} & \multicolumn{3}{c|}{BGL} & \multicolumn{3}{c|}{HDFS} & \multicolumn{3}{c|}{Thunderbirds} \\ \cline{2-10} 
 &
  \multicolumn{1}{l|}{F1 score} &
  \multicolumn{1}{l|}{Precision} &
  \multicolumn{1}{l|}{Recall} &
  \multicolumn{1}{l|}{F1 score} &
  \multicolumn{1}{l|}{Precision} &
  \multicolumn{1}{l|}{Recall} &
  \multicolumn{1}{l|}{F1 score} &
  \multicolumn{1}{l|}{Precision} &
  \multicolumn{1}{l|}{Recall} \\ \hline
\multicolumn{1}{|l|}{LogLLaMA w/o RL} &
  \multicolumn{1}{c|}{0.95723} &
  \multicolumn{1}{c|}{0.92236} &
  0.99484 &
  \multicolumn{1}{c|}{0.87867} &
  \multicolumn{1}{c|}{0.86755} &
  0.89007 &
  \multicolumn{1}{c|}{0.92517} &
  \multicolumn{1}{c|}{0.86293} &
  0.99709 \\ \hline
LogLLaMA &
  \multicolumn{1}{c|}{0.95927} &
  \multicolumn{1}{c|}{0.92748} &
  0.99333 &
  \multicolumn{1}{c|}{0.89363} &
  \multicolumn{1}{c|}{0.93897} &
  0.85248 &
  \multicolumn{1}{c|}{0.97407} &
  \multicolumn{1}{c|}{0.95679} &
  0.99198 \\ \hline
\end{tabular}%
}   

\end{table*}

\section{Experiments}

\subsection{Experimental Setup}

\textbf{Datasets.} We evaluate the overall performance of the baselines and out model on three publicly available datasets: BGL dataset, Thunderbird dataset, and HDFS dataset. Below is more detailed information on these three datasets.
\begin{itemize}
    \item BGL (Blue Gene/L Supercomputer System)\cite{BGL&Thunderbird_dataset}: The
    BGL dataset was generated from a Blue Gene/L supercomputer
    system at the Lawrence Livermore National Labs (LLNL). Each log message records information about hardware and software activities, error messages, warnings, informational messages, and other system events. The log contains alert and non-alert messages identified by alert category tags. Alert messages are considered as anomalies. The dataset has 4,747,963 log messages in total with 348,460 anomalous messages. We use sliding window = 60 seconds to group the log keys into log sequences.
    \item Thunderbird \cite{BGL&Thunderbird_dataset}: This dataset is collected from another supercomputer system at Sandia National Labs (SNL) in Albuquerque. We used the first 20,000,000 log messages from the original Thunderbird dataset with 758,562 anomalous messages. Like BGL dataset, we use sliding window = 60 seconds to group the log keys into log sequences.
    \item HDFS (Hadoop Distributed File System) \cite{HDFS_dataset&PCA}: The HDFS dataset was generated by logging events from the Hadoop Distributed File System (HDFS) in a real production environment. This dataset was collected over several weeks and represents a range of typical and anomalous operations within the Hadoop ecosystem. Anomaly labels were manually added by experts or generated based on error messages. The dataset has 11,172,157 log messages in total with 284,818 anomalous messages. The log messages are grouped by session IDs.
\end{itemize}

\textbf{Baselines.} We compare LogLLaMA with a variety of baseline methods whose source code is publicly available for us to implement. These baselines include both traditional machine learning methods and deep-learning methods:
\begin{itemize}
    \item PCA (Principal Component Analysis) \cite{HDFS_dataset&PCA}: This is a statistical technique used to simplify data while retaining its essential patterns. It reduces the dimensionality of data by transforming it into a new set of variables (called principal components) that are linear combinations of the original variables.
    \item iForest (Isolation Forest) \cite{iForest}: This is an unsupervised anomaly detection algorithm. Unlike traditional methods that profile normal data points and then detect anomalies based on deviation from this profile, iForest isolates anomalies directly by building a series of random trees.
    \item OCSVM (One-Class Support Vector Machine) \cite{OCSVM}: This is an unsupervised machine learning algorithm primarily used for anomaly detection. OCSVM is trained on only normal data and aims to learn a decision boundary that best separates the normal data points from the origin, which represents outliers or anomalies.   
    \item LogCluster \cite{LogCluster}: This is a density-based log clustering algorithm designed for grouping similar log messages in large-scale system logs. The method focuses on clustering log entries that exhibit similar patterns or behaviors, making it easier to identify normal operational trends. 
    \item DeepLog \cite{DeepLog}: This is a deep learning-based anomaly detection technique specifically designed for identifying unusual patterns in log data. The core idea behind DeepLog is to leverage Long Short-Term Memory (LSTM) networks to model the sequential nature of log entries, which are often generated in time series or event sequences.
    \item LogAnomaly \cite{LogAnomaly}: This is an anomaly detection approach that treats log streams as natural language sequences to identify both sequential and quantitative anomalies in system logs. By modeling logs in a way similar to language processing tasks, LogAnomaly can capture both the structure of log entries over time (sequential patterns) and unexpected behaviors or values (quantitative deviations).
    \item LogBERT \cite{LogBert}: This is a deep learning model based on BERT (Bidirectional Encoder Representations from Transformers), designed specifically for anomaly detection in log data. It utilizes a log language model to capture the patterns and dependencies present in normal log sequences. By adapting BERT’s architecture, which has proven effective for natural language processing tasks, LogBERT is able to understand the structure and relationships in log data.
 
\end{itemize}
\textbf{Evaluation metrics.} As the testing data is unbalanced in our datasets, we count the TP, FP, and FN values and calculate three metrics to evaluate our method and the baseline models, which can be calculated as the following:
\[
\text{Precision} = \frac{TP}{TP + FP}
\]
\[
\text{Recall} = \frac{TP}{TP + FN}
\]
\[
\text{F1 Score} = 2 \times \frac{\text{Precision} \times \text{Recall}}{\text{Precision} + \text{Recall}}
\]

\textbf{Implementation Details.} For the baseline models, we utilize the Loglizer\cite{Loglizer} package to evaluate PCA, OCSVM, iForest, and LogCluster for anomaly detection. DeepLog and LogAnomaly are evaluated using the Deep-loglizer\cite{Deep_loglizer} package. For LogBert, we use the opensource code provided by the authors to evaluate its performance.

For LogLLaMA, we utilize log parser Drain\cite{Drain} and regular expression to extract log keys with fixed lengths. Then we encode the log keys into tensor vectors as input data to the model. We initialize the LLaMA-2 foundation model with 18 layers, 12 attention heads. The input feature vector for each token has a dimension of 60. Because each dataset has a different number of log keys, we set the default K in the Top-K to be 50\% of the normal training log keys. It means during testing, if a log key is not in the top 50\% of the prediction list generated by our model, the log sequence is considered as anomaly. We use 10\% of the normal log sequences for training and the rest for testing. We train the foundation model for 100 epochs to achieve the best performance within reasonable time. During model fine-tuning, we use reinforcement learning and train the model for 10 episodes.

\subsection{Experimental Results}

\textbf{Performance on Log Anomaly Detection.} Table \ref{experimental results for all models} summarizes the performance of LogLLaMA and various baselines on HDFS, BGL, and Thunderbird datasets.

Our framework LogLLaMA achieves the highest F1 scores on three datasets with large margins by comparing with baselines with the highest F1 scores. It indicates that by using transformers and fine-tuning the foundation model with our downstream task, LogLLaMA can successfully understand the pattern of normal log sequences and capture anomalous sequences with high accuracy. Meanwhile, reinforcement learning can enhance the model's ability to identify shared patterns in normal log sequences, thereby improving its accuracy in predicting the next log key.

For the traditional machine learning methods, we observe that PCA, iForest, and OCSVM perform poorly on all of the datasets, indicating by their low F1 scores. Although PCA has a high recall on Thunderbird dataset, its poor precision lowers the F1 score significantly, which means the model does not balance between precision and recall well. LogCluster has a higher F1 score on BGL and HDFS datasets than PCA, iForest, and OCSVM. But it has a lower F1 score on Thunderbird dataset than that of OCSVM. We observe that LogCluster achieves high precisions but lower recall. As shown in Table \ref{experimental results for all models}, models PCA, iForest, OCSVM, and LogCluster can not balance between recall and precision.

For deep learning methods, DeepLog, LogAnomaly, and LogBert all have a higher F1 score on three datasets than the traditional machine learning methods. This suggests deep learning methods have greater potential for log anomaly detection than traditional machine learning methods. However, LogAnomaly has a low recall and a high precision.

\textbf{Ablation Studies.} To assess the impact of reinforcement learning on LogLLaMA's performance, we performed an ablation study comparing the results of LogLLaMA with and without the RL component. The results are summarized in Table \ref{ablation study table}. We notice that on all of the three datasets, LogLLaMA has achieved higher F1 scores than LogLLaMA without the RL component, which indicates that the RL component enhances the overall performance of LogLLaMA on detecting anomalous log patterns. Especially on the Thunderbird dataset, F1 score and precision have significantly improved after incorporating the RL component with only a small sacrifice on recall. We also note that even without RL, our model outperforms all of the baselines on BGL and HDFS dataset. With the RL component, our model outperforms all of the baselines on three datasets.

\section{conclusion}
Log anomaly detection is crucial for system security. Using Transformer-based LLMs is becoming a popular strategy for understanding complicated log messages and thus improve the accuracy of anomaly detection. In this work, we proposed LogLLaMA, a novel framework built upon LLaMA-2, for capturing anomalous log messages. We first finetune the model on normal log messages from three datasets for it to gain the ability to generating subsequent normal log messages. Then through reinforcement learning, we further improved the performance of the model for log anomaly detection. Experiments conducted across three large-scale datasets have proved the effectiveness of LogLLaMA, indicating significant improvements over existing state-of-the-art methods.

\bibliographystyle{IEEEtran}
% Generated by IEEEtran.bst, version: 1.14 (2015/08/26)

\end{document}